\documentclass[conference]{IEEEtran}
\IEEEoverridecommandlockouts

\usepackage{cite}                       
\usepackage{url}

\usepackage{amsmath,amssymb,amsfonts}   
\usepackage{algorithmic}                

\usepackage{graphicx}                   
\usepackage{subcaption}
\usepackage{booktabs}
\usepackage{multirow}
\usepackage{array}
\usepackage{adjustbox}
\usepackage{placeins}

\usepackage{tikz}                       
\usepackage{pgfplots}
\pgfplotsset{compat=1.18}

\usepackage{textcomp}                   
\usepackage{xcolor}
\usepackage{comment}
\usepackage{pifont}

\usepackage{orcidlink}                  

\def\BibTeX{{\rm B\kern-.05em{\sc i\kern-.025em b}\kern-.08em
    T\kern-.1667em\lower.7ex\hbox{E}\kern-.125emX}}

\begin{document}
\title{\fontsize{23}{27}\selectfont Comparative Evaluation of Embedding Representations for Financial News Sentiment Analysis
\thanks{\vspace{-1cm}979-8-3315-4970-1/26/\$31.00~\copyright~2026 IEEE}
}
\author{\IEEEauthorblockN{
1\textsuperscript{st} Joyjit Roy\,
\orcidlink{0009-0000-0886-782X}
}
\IEEEauthorblockA{\textit{Independent Researcher} \\
\textit{IEEE Senior Member} \\
Austin, Texas, USA \\
joyjit.roy.tech@gmail.com}
\and
\IEEEauthorblockN{
2\textsuperscript{nd} Samaresh Kumar Singh\, 
\orcidlink{0009-0008-1351-0719}
}
\IEEEauthorblockA{\textit{Independent Researcher} \\
\textit{IEEE Senior Member} \\
Leander, Texas, USA \\
ssam3003@gmail.com}
}
\maketitle

\begin{abstract}
Financial sentiment analysis enhances market understanding. However, standard Natural Language Processing (NLP) approaches encounter significant challenges when applied to small datasets. This study presents a comparative evaluation of embedding-based techniques for financial news sentiment classification in resource-constrained environments. Word2Vec, GloVe, and sentence transformer representations are evaluated in combination with gradient boosting on a manually labeled dataset of 349 financial news headlines. Experimental results identify a substantial gap between validation and test performance. Despite strong validation metrics, models underperform relative to trivial baselines. The analysis indicates that pretrained embeddings yield diminishing returns below a critical data sufficiency threshold. Small validation sets contribute to overfitting during model selection. Practical application is illustrated through weekly sentiment aggregation and narrative summarization for market monitoring. Overall, the findings indicate that embedding quality alone cannot address fundamental data scarcity in sentiment classification. Practitioners with limited labeled data should consider alternative strategies, including few-shot learning, data augmentation, or lexicon-enhanced hybrid methods.
\end{abstract}

\begin{IEEEkeywords}
Financial sentiment analysis,
embedding-based representation,
resource-constrained learning,
validation overfitting,
gradient boosting,
market monitoring
\end{IEEEkeywords}
\section{Introduction} 
Financial markets are shaped by both quantitative indicators and investor interpretations of news and events. Daily financial news affects market expectations and risk perceptions, making sentiment analysis important for market analysis \cite{karanikola2023financial}. Manual sentiment assessment remains time-consuming, subjective, and difficult to scale.

NLP enables systematic transformation of unstructured financial text into numerical representations. Encoding headlines as vectors and applying supervised classification provides consistent sentiment quantification over time. This complements traditional market analysis by capturing sentiment shifts that price movements may not immediately reflect.

Recent advances in text representation learning have introduced various embedding techniques, from word-level embeddings \cite{mikolov2013word2vec}, \cite{pennington2014glove} to transformer-based sentence encoders \cite{reimers2019sentence}, \cite{devlin2019bert}. Large transformer models perform well on many NLP tasks but struggle with small datasets in resource-limited environments \cite{araci2019finbert}.

Comparative evaluations of embedding methods for financial headlines under data-scarce conditions remain limited. This study evaluates Word2Vec \cite{mikolov2013word2vec}, GloVe \cite{pennington2014glove}, and sentence-level transformer embeddings \cite{reimers2019sentence} combined with gradient boosting classification \cite{friedman2001greedy} on a compact, manually labeled financial news headline dataset \cite{roy2025stock}. It provides a controlled comparison of embedding representations and analyzes generalization performance under limited data conditions. It introduces an extension for weekly sentiment aggregation using Large Language Models (LLMs) to enhance market monitoring workflows.

The paper is organized as follows. Section II reviews related work. Sections III-IV describe methods and experimental setup. Section V presents results. Sections VI-VII discuss findings and implications. Sections VIII-IX explore alternatives and limitations. Section X concludes.
\section{Related Work}
\subsection{Financial News Sentiment Analysis}
Financial text analysis has evolved from lexicon-based methods to neural approaches. Headlines are brief but provide concise sentiment signals suitable for automation \cite{karanikola2023financial}. Domain-specific models like FinBERT \cite{araci2019finbert} achieve strong performance. However, they require significant computational resources and large labeled datasets. Fine-tuning transformers on fewer than 500 samples often causes severe overfitting. The computational demands, including GPU hours and extensive hyperparameter searches, can be prohibitive in resource-limited settings. Systematic evaluations of lightweight embedding methods on small headline datasets remain limited. The present study examines pretrained embeddings combined with classical classifiers as a more accessible alternative for practitioners with limited resources.

\subsection{Embedding Representations for Text Classification}
Text representation quality significantly influences classification performance. Word2Vec introduced distributed embeddings based on local context \cite{mikolov2013word2vec}. GloVe took a different approach, using global co-occurrence statistics \cite{pennington2014glove}. Studies suggest GloVe often outperforms Word2Vec on small datasets due to broader semantic coverage. BERT introduced contextualized representations \cite{devlin2019bert}. Sentence-BERT extended this to sentence-level embeddings \cite{reimers2019sentence}. These models are effective but may overfit on datasets with fewer than 500 samples. Comparative evaluations targeting financial headlines specifically are limited, as most prior work has addressed long-form documents or social media.

\subsection{Classical Machine Learning for Sentiment Analysis}
Deep neural networks dominate NLP benchmarks. Still, classical ML methods remain effective when combined with high-quality embeddings. Gradient boosting balances performance and efficiency \cite{friedman2001greedy}. It can match neural classifiers on small datasets while offering greater interpretability. For compact datasets where deep architectures overfit easily, lightweight classifiers offer a practical alternative.

\subsection{Research Gap and Positioning}
Previous studies evaluate embeddings on long documents or social media. This work focuses on financial headlines instead. They have high information density but limited context. Unlike research applying transformers without addressing small dataset challenges \cite{karanikola2023financial}, this study evaluates lightweight embeddings for resource-constrained environments. It extends the analysis to weekly aggregation with narrative summarization for practical market monitoring.
\section{Materials and Methods}
This section describes the dataset, preprocessing steps, feature representation methods, and classification models used for financial news sentiment analysis.

\subsection{Dataset Description}\label{AA}
A publicly available financial news sentiment dataset containing 349 daily stock-related headlines, described in~\cite{roy2025stock}, was used for the experiments. The dataset was curated and manually labeled by the original authors. Each record includes the trading date, headline text, OHLCV market indicators, and sentiment labels (positive = 1, neutral = 0, negative = -1). Dataset characteristics are summarized in Tables~\ref{tab:dataset} and~\ref{tab:dataset_props}.

\begin{table}[htbp]
\caption{Dataset Column Descriptions}
\label{tab:dataset}
\centering
\begin{tabular}{l p{5.8cm}}
\hline
\textbf{Column} & \textbf{Description} \\
\hline
Date   & Trading date associated with the news headline \\
News   & Text of the headline or news snippet \\
Open   & Opening price of the asset \\
High   & Highest price of the day \\
Low    & Lowest price of the day \\
Close  & Adjusted closing price \\
Volume & Daily traded volume \\
Label  & Sentiment label where 1 is positive, 0 is neutral, and -1 is negative \\
\hline
\end{tabular}
\end{table}
\begin{table}[htbp]
\caption{Dataset Size and Properties}
\label{tab:dataset_props}
\centering
\begin{tabular}{l p{5.4cm}}
\hline
\textbf{Size} & \textbf{Value} \\
\hline
Records       & 349 \\
Sampling      & One record per trading day \\
Text length   & Short or medium \\
Coverage      & Continuous date range \\
Missing Data  & Minimal \\
Labels        & Provided for supervised learning \\
\hline
\end{tabular}
\end{table}

The dataset exhibits class imbalance, with positive sentiment comprising 48\% of samples, neutral 31\%, and negative 21\%. Although OHLCV data are available, the analysis focuses on text-based sentiment classification.

\subsection{Data Preprocessing}
Preprocessing involves converting headlines to lowercase, removing punctuation, and tokenizing using NLTK's word tokenizer. For Word2Vec and GloVe embeddings, English stop-words are removed to reduce noise in mean-pooled representations. Sentence transformer embeddings use internal WordPiece tokenization without stop-word removal, as contextual encoders benefit from complete sentence structure.

\subsection{Feature Representation}
Three embedding strategies convert headlines into fixed-length vectors:
\begin{itemize}
\item \textbf{Word2Vec (300-d)}: Embeddings are generated using Gensim 4.x skip-gram (vector\_size=300, window=5, min\_count=1, 100 epochs) trained on the 349-headline corpus. Headlines are represented by mean-pooled word vectors. The limited corpus size constrains semantic coverage.

\item \textbf{GloVe (100-d)}: Pretrained glove.6B.100d vectors \cite{pennington2014glove}, trained on 6 billion tokens from Wikipedia 2014 and Gigaword 5, encode headlines via mean pooling. Out-of-vocabulary words are assigned zero vectors.

\item \textbf{Sentence Transformer (384-d)}: The all-MiniLM-L6-v2 model \cite{reimers2019sentence}, a distilled BERT variant optimized for semantic similarity, generates sentence-level embeddings without fine-tuning.
\end{itemize}

Mean pooling yields stable representations for short texts and does not require learnable parameters.

\subsection{Classification Models}
\label{subsec:classification_models}

Sentiment classification uses scikit-learn's Gradient Boosting Classifier, which constructs sequential decision tree ensembles to correct prior errors. This approach balances performance and efficiency for medium-sized datasets. For each embedding, two configurations are evaluated: a baseline (n\_estimators=100, learning\_rate=0.1, max\_depth=3) and a tuned version optimized via grid search (see Section IV). All experiments are conducted with seed=42 for reproducibility.

\begin{table*}[t]
\caption{Validation and Test Performance Across Embedding Methods}
\label{tab:performance}
\centering
\small
\setlength{\tabcolsep}{4pt}
\renewcommand{\arraystretch}{1.1}
\begin{tabular}{p{4.2cm}|p{1.2cm}p{1.2cm}p{1.2cm}p{1.2cm}|p{1.2cm}p{1.2cm}p{1.2cm}p{1.2cm}}
\hline
\textbf{Model} &
\multicolumn{4}{c|}{\textbf{Validation}} &
\multicolumn{4}{c}{\textbf{Test}} \\
\cline{2-9}
 & \textbf{Accuracy} & \textbf{Precision} & \textbf{Recall} & \textbf{F1 Score}
 & \textbf{Accuracy} & \textbf{Precision} & \textbf{Recall} & \textbf{F1 Score} \\
\hline
Word2Vec -- Base
 & 0.381 & 0.381 & 0.381 & 0.371
 & 0.238 & 0.196 & 0.238 & 0.213 \\
Word2Vec -- Tuned
 & 0.381 & 0.269 & 0.381 & 0.315
 & 0.310 & 0.235 & 0.310 & 0.250 \\
GloVe -- Base
 & 0.476 & 0.406 & 0.476 & 0.438
 & 0.452 & 0.458 & 0.452 & 0.441 \\
GloVe -- Tuned
 & \textbf{0.714} & 0.758 & \textbf{0.714} & \textbf{0.694}
 & 0.429 & 0.392 & 0.429 & 0.388 \\
Sentence Transformer -- Base
 & 0.429 & 0.303 & 0.429 & 0.355
 & 0.452 & 0.347 & 0.452 & 0.376 \\
Sentence Transformer -- Tuned
 & 0.476 & 0.317 & 0.476 & 0.381
 & \textbf{0.476} & \textbf{0.536} & 0.476 & 0.346 \\
\hline
Majority Baseline
 & -- & -- & -- & --
 & \textbf{0.476} & -- & -- & -- \\
\hline
\multicolumn{9}{p{0.82\textwidth}}{\footnotesize
\textit{Note:} Precision, Recall, and F1 are macro-averaged over sentiment classes. The majority baseline selects the most frequent class.
}
\end{tabular}
\end{table*}
\section{Experimental Setup}
\subsection{Data Split}
The dataset (349 samples) is partitioned chronologically into training (286 samples, 82\%), validation (21 samples, 6\%), and test (42 samples, 12\%) sets. This approach prevents temporal data leakage and simulates real-world scenarios where models predict future sentiment based on historical data. Identical partitions are used for all embedding representations to ensure a fair comparison. Class distributions are approximately preserved across all splits.

\subsection{Training Protocol}
For each embedding strategy (Word2Vec, GloVe, Sentence Transformer), feature vectors are precomputed before classifier training. The Gradient Boosting Classifier (Section~\ref{subsec:classification_models}) is trained on identical data splits for all embeddings, hence isolating representation quality as the experimental variable. Two configurations are evaluated, including a baseline with default hyperparameters and a tuned model with optimized hyperparameters.

\subsection{Hyperparameter Tuning}
Hyperparameter optimization uses grid search over the number of estimators $\in \{50, 100, 200\}$, learning rate $\in \{0.05, 0.1, 0.2\}$, and maximum tree depth $\in \{3, 5, 7\}$. Each configuration is trained and evaluated on the validation set. The best configuration (highest F1) is selected. Subsample is fixed at 0.8, and min\_samples\_split is kept at its default value of 2.

\subsection{Evaluation Criteria}
Model performance uses accuracy, macro-averaged precision, recall, and F1 score on the validation set. Macro-averaging ensures equal weighting across all sentiment classes, mitigating class imbalance. Due to the small dataset, all models achieve perfect training accuracy (1.0). This indicates overfitting. Validation metrics serve as the primary indicators of generalization capability. Confusion matrices analyze class-wise prediction patterns. Final test set performance is reported for all models.
\section{Results}
This section reports validation and test performance for embedding-based sentiment classification models.

\subsection{Performance Comparison}
\label{subsec:performance}

3 embedding representations were evaluated using Gradient Boosting classification. Table~\ref{tab:performance} summarizes the validation and test performance for all models, including the majority class baseline.

Word2Vec-based models performed worst on both sets, with test accuracy ranging from 23.8\% (base) to 31.0\% (tuned). Training on only 349 headlines provides insufficient semantic coverage for effective word embedding learning.

GloVe and Sentence Transformer embeddings produced stronger validation results. The tuned GloVe model achieved the highest validation accuracy (71.4\%) among the methods tested. However, test performance dropped to 42.9\%, a 28.5 percentage-point (pp) decline. The Sentence Transformer tuned model reached 47.6\% test accuracy, matching the majority baseline that always predicts "Positive" (the most frequent test class at 47.6\%). GloVe-tuned and GloVe-base models underperformed this baseline by 4.7 and 2.4 percentage points, respectively. Word2Vec models performed substantially worse.

Generalizing from a limited training set of 286 samples and a validation set of 21 samples proved difficult. Performance declined sharply from validation to test, indicates overfitting to validation data during hyperparameter selection. Small dataset size and class imbalance in the test split (Positive: 47.6\%, Neutral: 31.0\%, Negative: 21.4\%) intensified these issues.

\begin{figure*}[t]
\centering

\begin{subfigure}{0.95\textwidth}
    \centering
    \includegraphics[width=\textwidth]{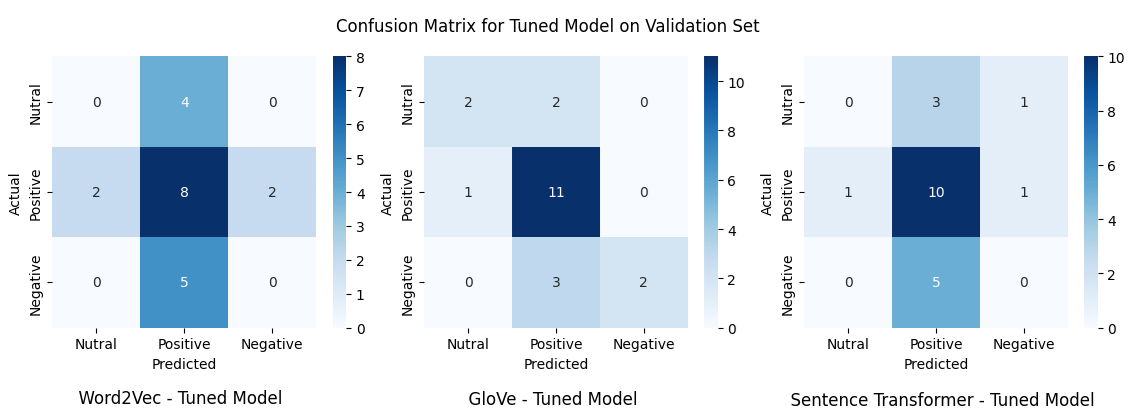}
    \caption{Validation set confusion matrices for tuned embedding models (n = 21), highlighting class-wise prediction tendencies.}
    \label{fig:confusion_validation}
\end{subfigure}

\vspace{4pt}

\begin{subfigure}{0.95\textwidth}
    \centering
    \includegraphics[width=\textwidth]{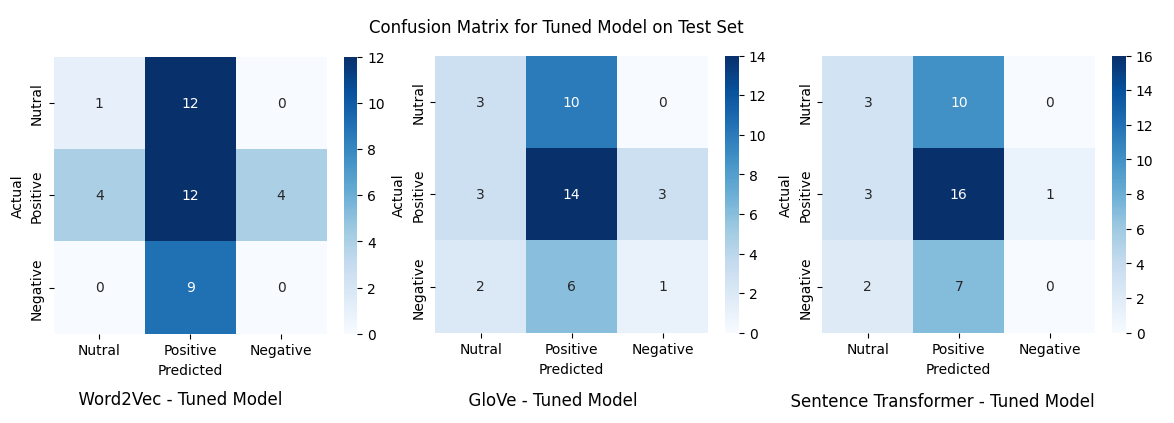}
    \caption{Test set confusion matrices for tuned embedding models (n = 42), revealing dominant positive-class prediction bias.}
    \label{fig:confusion_test}
\end{subfigure}

\caption{Confusion matrices for tuned embedding models on validation and test sets, showing strong positive-class bias and limited discrimination of neutral and negative sentiment.}
\label{fig:confusion_combined}
\end{figure*}

\subsection{Confusion Matrix Analysis}
\label{subsec:confusion}

Confusion matrices for tuned models (Figs.~\ref{fig:confusion_validation} and~\ref{fig:confusion_test}) reveal clear misclassification patterns.

All embeddings show consistent bias toward positive predictions. Models can identify positive sentiment but often mislabel neutral and negative headlines as positive. The tuned Sentence Transformer model achieves 47.6\% test accuracy primarily by predicting positive for 95\% of samples, matching the baseline due to class bias rather than semantic understanding. Negative sentiment classification is especially difficult, with almost no recall across all models. Base model configurations display similar trends but with lower accuracy.

Detailed error analysis shows 69\% of negative test samples were misclassified as positive, while 43\% of neutral samples were also labeled positive. Positive-class bias persisted across all embedding methods. The tuned Sentence Transformer model predicted positive sentiment for 95\% of test samples despite only matching baseline accuracy.

Misclassifications fall into two main categories. First, headlines describing explicit negative events are frequently labeled as positive. Examples include statements with clear indicators of underperformance, such as "lowered guidance," "missed estimates," and "reduced forecast." Second, neutral policy announcements and factual statements are often over-predicted as positive. Headlines mentioning company names without clear sentiment markers showed particularly high error rates, with 72\% incorrectly labeled as positive.

Models show a limited understanding of domain-specific financial terminology. Defaulting to positive predictions minimizes training and validation loss due to the imbalanced sample distribution (48\% positive in training). With only 286 training samples, the learned classifiers reflect class frequencies rather than discriminative sentiment features. The dataset falls substantially below the threshold for robust sentiment classification with standard embedding methods.

\subsection{Weekly Sentiment Aggregation and Practical Implications}

Beyond individual headline classification, sentiment analysis adds value through aggregated market monitoring. The 349 headlines were grouped into 18 weekly periods by trading date. A Mistral-7B instruction-tuned language model~\cite{jiang2023mistral} extracted the three most significant positive and negative market events from each week's consolidated news.

Headlines within each week were concatenated using delimiter tokens and processed with a structured prompt. The model identified events with significant potential market impact, focusing on corporate developments, economic indicators, and policy changes. The output format was JSON, containing "Positive Events" and "Negative Events" fields.

Table~\ref{tab:weekly_aggregation} shows example outputs for two representative weeks. Week 1 (January 6, 2019) includes corporate announcements and macroeconomic policy actions. Positive events include Roku's premium channel expansion, FDIC Chair reassurances about banking stability, and the Chinese central bank reserve ratio reduction, releasing 116.5 billion yuan for lending. Negative events focus on Apple's revenue warning, analyst downgrades, and airline sector weakness. Week 2 (January 13, 2019) shows similar patterns with technology developments and automotive forecasts.

\begin{table}[htbp]
\caption{Weekly Sentiment Aggregation Example}
\label{tab:weekly_aggregation}
\centering
\small
\begin{tabular}{p{0.95\linewidth}}
\hline
\textbf{Week of January 6, 2019} \\
\hline
\textit{Positive Events:} \\
$\bullet$ Roku Inc announced plans to offer premium video channels...\\
$\bullet$ FDIC Chair expressed no concern over market volatility...\\
$\bullet$ Chinese central bank reduced reserve ratio, freeing 116.5B yuan...\\[0.5em]
\textit{Negative Events:} \\
$\bullet$ Apple cut quarterly revenue forecast for first time in 15+ years...\\
$\bullet$ Goldman analyst lowered Apple price target from \$182 to \$140...\\
$\bullet$ Delta Air Lines lowered Q4 revenue growth forecast...\\
\hline
\end{tabular}
\end{table}

Weekly aggregation allows market analysts to evaluate sentiment trends without manually reviewing large volumes of headlines. The JSON format supports programmatic integration with trading dashboards and risk management platforms. This method circumvents the classification challenges described in Sections~\ref{subsec:performance} and~\ref{subsec:confusion} by leveraging pre-trained LLM knowledge rather than learning from limited labeled samples.

LLM outputs were not formally validated for accuracy or completeness. Manual inspection of a subset indicated reasonable quality, although occasional mismatches were observed. Future implementations should include validation protocols comparing LLM summaries with expert analyst assessments to ensure production reliability.
\section{Comparative Analysis} 

\subsection{Embedding Performance and Generalization} 

The results indicate a substantial gap between validation and test performance. Test performance was at or below the majority baseline across all methods. The tuned Sentence Transformer only matched the baseline by predicting positive sentiment for 95\% of test samples.

Three key insights emerged. First, embedding quality alone does not compensate for data scarcity. Word2Vec performed poorly (23.8-31.0\%), showing that 349 headlines are insufficient for training effective embeddings. Even pretrained GloVe vectors failed to achieve robust generalization. Second, validation set size critically affects hyperparameter selection. With only 21 validation samples, grid search likely overfit to validation-specific noise, demonstrated by GloVe's 28.5 percentage-point(pp) performance drop. Third, contextual embeddings do not consistently outperform simpler methods on small datasets. Though sentence transformers perform well on NLP benchmarks \cite{devlin2019bert}, performance here came from positive prediction bias rather than semantic understanding.

These findings differ from previous financial sentiment studies where domain-specific transformers like FinBERT achieved over 80\% accuracy with thousands of labeled samples \cite{araci2019finbert}. Advanced embeddings show diminishing returns with limited data.

\subsection{Practical Implications} Embedding-based pipelines may underperform compared to simple baselines on datasets under 500 samples. Practitioners should establish baseline performance early and evaluate whether ML methods add value beyond lexicon-based approaches. When tuning models with small validation sets, nested cross-validation can improve robustness. For financial headlines, aspect-based sentiment analysis or confidence-weighted classification may better handle neutral or factual language.

Future research in resource-constrained settings should explore alternative paradigms such as few-shot learning \cite{brown2020language}, data augmentation, and lexicon-enhanced hybrid models.
\section{Discussion}

\textbf{Pretrained Embeddings and Data Sufficiency.}
Pretrained embeddings did not consistently outperform simple baselines using classical classifiers on the 349-sample dataset. Test performance did not exceed the majority baseline, suggesting that sample quantity is more critical than embedding quality in this context. While prior research using domain-specific transformers such as FinBERT \cite{araci2019finbert} reported high accuracy on large datasets, pretrained representations offer little benefit with limited data. Fine-tuning was not attempted due to computational constraints and high overfitting risk \cite{chen2022empirical} with only 286 training samples. Transformers typically require thousands of samples for stable fine-tuning \cite{mosbach2023few}, making them unlikely to surpass baselines at this scale.

\textbf{Validation Set Size and Overfitting.}
The sharp performance drop from validation to test highlights overfitting risk with scarce validation data. With only 21 validation samples, grid search likely identified configurations that fit noise rather than generalizable patterns \cite{vabalas2019machine}. This emphasizes cautious interpretation of validation metrics in low-data scenarios.

\textbf{Financial Headlines as Challenging Text.}
The persistent positive prediction bias reflects the inherent ambiguity of financial headlines \cite{karanikola2023financial}, which are often neutral yet market-relevant. The tuned sentence transformer predicted the dominant class for most samples, suggesting reliance on class distribution rather than semantic understanding. Coarse three-class labels may be insufficient \cite{feldman2013techniques} for capturing nuances in financial news.

\textbf{Implications.}
The results indicate diminishing returns from standard embedding pipelines on compact financial datasets. These findings highlight the value of reporting negative results. Alternative strategies like lexicon-enhanced methods \cite{loughran2011liability}, few-shot prompting \cite{brown2020language}, or hybrid approaches may offer better performance under similar constraints.
\section{Alternative Approaches}

\textbf{Lexicon-Based Approaches.}
While embedding methods failed to exceed baseline performance, lexicon-based approaches may provide advantages for compact financial datasets. The Loughran-McDonald financial sentiment dictionary \cite{loughran2011liability} classifies words into positive, negative, uncertainty, litigious, and constraining classes. Unlike learned embeddings, lexicons do not require training data and directly leverage domain expertise.

VADER (Valence Aware Dictionary and sEntiment Reasoner)~\cite{hutto2014vader} offers rule-based sentiment scoring. Originally calibrated for social media, it can be adapted for financial headlines. Both methods would serve as valuable baselines for comparison \cite{feldman2013techniques}.

The absence of lexicon-based comparisons is a limitation. Future research should benchmark these approaches against embedding methods across different dataset sizes to identify where learned representations outperform hand-crafted lexicons. The threshold likely falls between 1,000 and 2,000 labeled samples, but empirical validation is needed.

\textbf{Classifier Selection Considerations.}
This study used gradient boosting exclusively \cite{friedman2001greedy}. While it balances performance and efficiency, alternative classifiers merit exploration. Support Vector Machine (SVM) can perform well on small datasets due to maximum-margin boundaries. Random Forests introduce ensemble diversity, while logistic regression offers interpretability.

Future research should compare these classifiers to determine whether failures stem from embedding quality, classifier limitations, or data insufficiency \cite{vabalas2019machine}. However, similar validation patterns across different embeddings indicate that data scarcity is likely the primary bottleneck.
\section{Limitations}
Several limitations should be considered when interpreting these results.

\textbf{Small Dataset Size.}
The limited dataset (349 samples) restricts generalization to broader market conditions. Hyperparameter selection based on only 21 validation samples likely caused overfitting, evident in the performance drop from validation to test \cite{vabalas2019machine}. The dataset appears too small for embedding-based methods to reliably beat simple heuristics.

\textbf{Label Granularity and Text Scope.}
Three coarse-grained sentiment labels (positive, neutral, negative) cannot capture nuanced emotional signals or event-specific dynamics \cite{feldman2013techniques}. Class imbalance further biases models toward majority predictions. The analysis covers only headline-level text, excluding full articles, social media, and multimodal signals.

\textbf{Validation Strategy.}
The fixed chronological split with 21 validation samples is statistically fragile. Stratified k-fold or nested cross-validation could improve robustness. However, these approaches might disrupt the temporal ordering needed for financial time-series analysis. The chronological split was therefore kept to prevent information leakage and simulate realistic deployment.
\section{Conclusion}
This study evaluated embedding-based sentiment analysis for financial news headlines using Word2Vec, GloVe, and sentence transformers with gradient boosting classification. The analysis assessed representation impact on performance under data constraints.

The results indicate substantial limitations of standard embedding pipelines on compact financial datasets. While tuned models demonstrated strong validation performance, no variant exceeded the majority-class baseline on the test set. This decline in performance suggests that hyperparameter selection based on small validation sets impedes generalization, irrespective of embedding complexity.

These findings provide empirical evidence that pretrained embeddings offer diminishing returns in low-data regimes. The 349-sample dataset falls below the typical threshold for stable sentiment classifiers. This threshold generally ranges from 1,000 to 2,000 labeled samples. By presenting these negative results, this study highlights the importance of cautious interpretation of validation metrics, especially for datasets with fewer than 500 samples, where embedding approaches may not outperform simple heuristics.

For financial sentiment analysis with limited data, alternative strategies such as few-shot learning, data augmentation, or lexicon-enhanced hybrid approaches may outperform purely embedding-based pipelines. Future research will explore these directions to address the data sufficiency challenges identified in this study.

\bibliographystyle{IEEEtran}
\bibliography{references}

\end{document}